\documentclass[runningheads]{llncs}
\usepackage{hyperref}
\usepackage{amssymb}
\usepackage[T1]{fontenc}
\usepackage{multirow}
\usepackage{booktabs}
\usepackage{makecell}
\usepackage{bm}
\usepackage{graphicx,verbatim}
\begin{document}
\title{KP-INR: A Dual-Branch Implicit Neural Representation Model for Cardiac Cine MRI Reconstruction}
%

\author{Donghang Lyu\inst{1} \and Marius Staring\inst{1} \and Mariya Doneva\inst{4} \and Hildo J. Lamb\inst{1} \and Nicola Pezzotti\inst{2,3}}

\institute{Department of Radiology, Leiden University Medical Center, Leiden, The Netherlands\\ 
\email{d.lyu@lumc.nl}
\and
Cardiologs, Philips, Paris, France
\and
Faculty of Computer Science, Eindhoven University of Technology, Eindhoven, The Netherlands
\and
Philips Innovative Technologies, Hamburg, Germany}


\maketitle              
\begin{abstract}
Cardiac Magnetic Resonance (CMR) imaging is a non-invasive method for assessing cardiac structure, function, and blood flow. Cine MRI extends this by capturing heart motion, providing detailed insights into cardiac mechanics. To reduce scan time and breath-hold discomfort, fast acquisition techniques have been utilized at the cost of lowering image quality. Recently, Implicit Neural Representation (INR) methods have shown promise in unsupervised reconstruction by learning coordinate-to-value mappings from undersampled data, enabling high-quality image recovery. However, current existing INR methods primarily focus on using coordinate-based positional embeddings to learn the mapping, while overlooking the feature representations of the target point and its neighboring context. In this work, we propose KP-INR, a dual-branch INR method operating in k-space for cardiac cine MRI reconstruction: one branch processes the positional embedding of k-space coordinates, while the other learns from local multi-scale k-space feature representations at those coordinates. By enabling cross-branch interaction and approximating the target k-space values from both branches, KP-INR can achieve strong performance on challenging Cartesian k-space data. Experiments on the CMRxRecon2024 dataset confirms its improved performance over baseline models and highlights its potential in this field.
\keywords{Cardiac Cine MRI Reconstruction \and Implicit Neural Representation \and Dual-Branch \and Unsupervised Reconstruction}

\end{abstract}
\section{Introduction}

Cardiac cine Magnetic Resonance Imaging (MRI) is essential for evaluating cardiovascular function, offering exceptional soft tissue contrast and the ability to capture dynamic heart motion. However, long acquisition times and repeated breath-holds can cause patient discomfort and hinder high-resolution imaging. To address these issues, increasing acceleration factors to speed up data acquisition and reduce the number of breath-holds has gained great attention, which also poses challenges in effectively reconstructing highly accelerated data.

Parallel Imaging (PI)~\cite{1,3} and Compressed Sensing (CS)~\cite{2} are widely used for reconstructing undersampled k-space data, but achieving high spatial and temporal resolution remains challenging. In recent years, deep learning methods have shown strong potential in cardiac cine MRI reconstruction~\cite{4,5,6,7}, offering high-quality results at higher acceleration factors. However, most approaches rely on fully sampled data for supervision, which is costly and often requires multiple heartbeats, introducing potential inconsistencies. As a result, unsupervised reconstruction has emerged as a promising alternative to overcome these limitations.

INR methods are gaining attention in computer vision for modeling physical properties from spatial coordinates~\cite{8,9} and are being explored in many medical imaging tasks, such as super-resolution~\cite{10}, registration~\cite{11}, segmentation~\cite{12}, and reconstruction~\cite{13,14}. In MRI reconstruction, some INR methods~\cite{15,28} have shown effective scan-specific reconstructions at high acceleration factors without requiring fully sampled data. Cardiac cine MRI poses additional challenges due to its temporal dimension, but some methods~\cite{14,25,29} have also achieved promising results in this domain. Among them, NIK~\cite{14} is a representative work that employs an MLP network with inputs (time point, k-space coordinates, coil index) to approximate sampled k-space values, learning the full k-space distribution to estimate missing data. Another INR method~\cite{29} feeds all spatio-temporal coordinates of the image series into a hash grid and employs an MLP to predict the real and imaginary components. Regularizations are then applied in the image domain, along with constraints to ensure consistency with the acquired k-space data. However, these methods primarily utilize coordinate-based encoding, neglecting the incorporation of feature representations that capture both the coordinates and their surrounding spatial and temporal context, which can serve as a regularizer and motivate the learning with positional embeddings.


In this work, we propose \textbf{KP-INR}, a novel dual-branch INR model for cardiac cine MRI reconstruction. In cine MRI k-space, spatial and temporal dimensions exhibit local neighborhood structure~\cite{36} and this potential relationship can also be used for cine MRI reconstruction~\cite{50,51}. Additionally, direct coordinate-to-value mapping without regularization increases the risk of overfitting. Therefore, KP-INR adopts a dual-branch architecture: one branch follows the standard INR design by applying positional encoding to the input coordinates, enabling the modeling of high-frequency variations, while the other branch serves as a regularizer, whose input is k-space features of coordinates, obtained through a complex-valued U-Net used as an auto-encoder. These high-dimensional feature embeddings also include neighborhood information along spatial and temporal dimensions, enabled by the use of convolution and recurrent operations. Both branches aim to approximate sampled k-space values, with interactions between them enabling information exchange to enhance learning in each branch. Furthermore, the optimization of KP-INR is composed of two alternating stages: (1) an optimization stage where the model parameters are trained; (2) an inference stage, in which a Cartesian grid of all coordinates is fed into the frozen KP-INR to generate refined multi-coil k-space. This refined k-space is then passed to KP-INR to continue optimization, making it possible to obtain improved feature representations. The key contributions of KP-INR are as follows: 
\begin{enumerate}
    \item We propose a dual-branch INR framework that integrates multi-scale k-space features with positional embeddings of k-space coordinates. To the best of our knowledge, this is the first INR method to jointly leverage both k-space features and coordinate representations for cine MRI reconstruction.
    \item The overall optimization of KP-INR consists of two alternating stages: an optimization stage followed by an inference stage, which generates a refined multi-coil k-space to allow for improved k-space feature representations in the subsequent optimization.
    \item KP-INR is evaluated on the public CMRxRecon2024 dataset~\cite{24} across two acceleration factors (4$\times$, 8$\times$) and Cartesian sampling patterns (uniform and Gaussian), providing a comprehensive assessment and showing consistently improved overall reconstruction performance over baseline methods.
\end{enumerate}

\section{Methodology}
\subsection{KP-INR}
\begin{figure}[tb!]
\includegraphics[width=\textwidth]{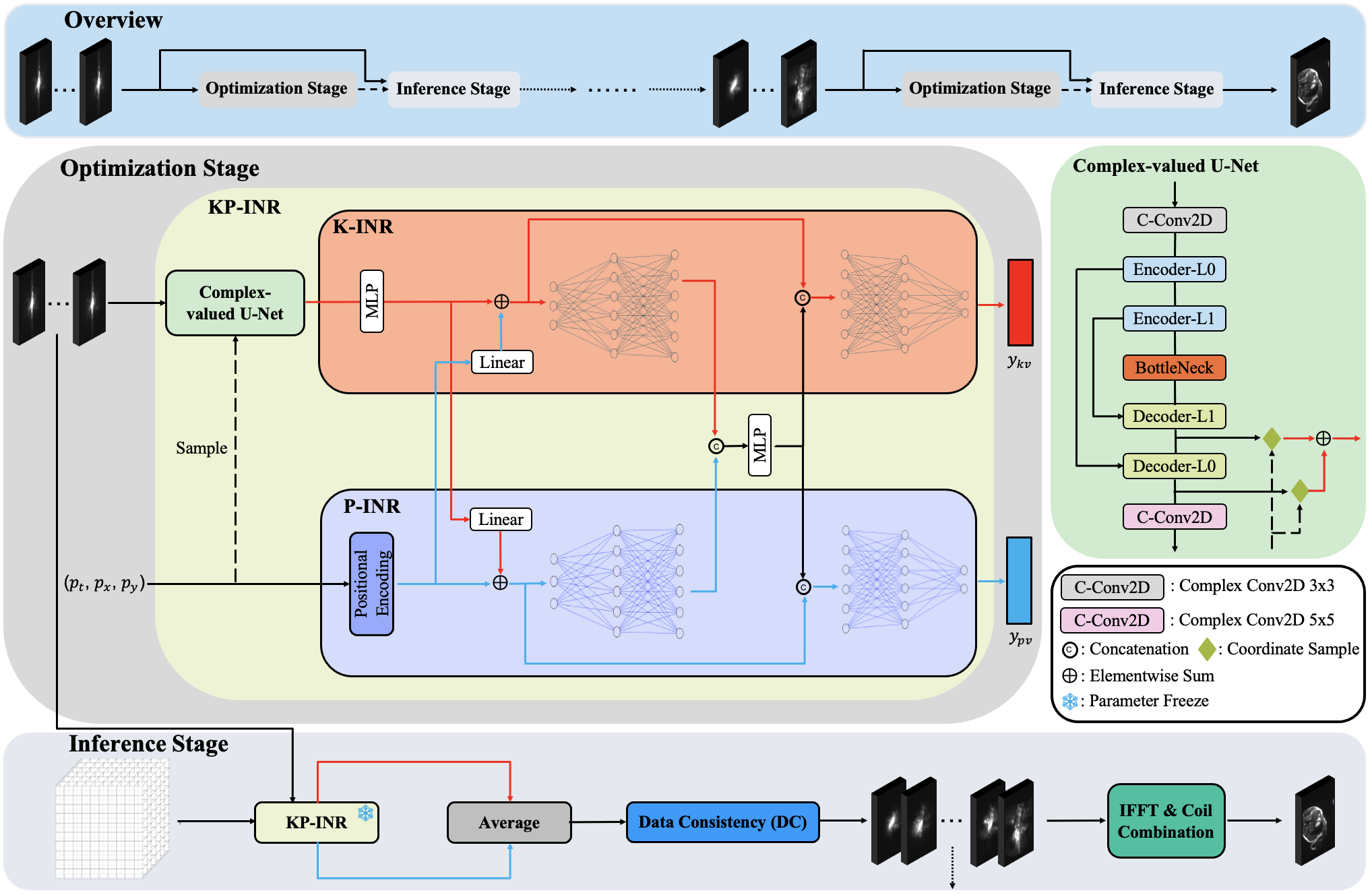}
\caption{The overall KP-INR architecture and its alternating two-stage optimization are illustrated. The top shows iterative k-space updates to the complex-valued U-Net of KP-INR, while the dashed lines indicate the optimization flow of KP-INR. Blue and red lines in KP-INR indicate positional embeddings and k-space feature flows, respectively. The bottom depicts the inference pipeline, and the right shows the complex-valued U-Net structure and k-space feature sampling process.} \label{fig1}
\end{figure}

Overall, the structure of KP-INR consists of three key components: a complex-valued U-Net, an INR branch with k-space feature input (K-INR), and an INR branch with positional embedding input (P-INR), as shown in Figure~\ref{fig1}. Globally, the model takes two inputs: multi-coil k-space and sampled coordinate set. Given an initial multi-coil undersampled k-space data $\mathbf{y} \in \mathbb{C}^{H \times W \times C \times T}$, where $H$, $W$, $C$, and $T$ represent the height, width, coil number, and frame number, respectively, it is fed into a complex-valued U-Net for auto-encoding. Additionally, the format of a k-space coordinate is $p_{i}=[p_{t}^{i},p_{x}^{i},p_{y}^{i}]^{T}$, with $p_{t}^{i}$ the frame index and $p_{x}^{i}$, $p_{y}^{i}$ the spatial position in k-space. Here, $i=1,2,\ldots,N$ and $N$ is the total number of sampled coordinates. Then the coordinate set $P \in \mathbb{R}^{N \times 3}$ is not only used as input to the P-INR branch but also to sample k-space feature embeddings from the decoder part of complex-valued U-Net. The K-INR branch receives the k-space embedding $k_{f} \in \mathbb{R}^{N \times L}$ derived from $P$, where $L$ is the embedding length, and outputs $\bm{y}_{kv} \in \mathbb{R}^{N \times 2C}$, while the P-INR branch uses the positional embedding of the coordinate set $P$ to output $\bm{y}_{pv} \in \mathbb{R}^{N \times 2C}$. Each output corresponds to the predicted real and imaginary values of all coils at the given coordinates. During the optimization stage, both branches are optimized to approximate the acquired multi-coil k-space values $\bm{y}$, with each branch $G_{\theta}$ trained as follows:
\begin{equation}
    \theta^* = \arg\min_\theta \|G_\theta(P) - \bm{y}\|_2.
\end{equation}

\subsubsection{Complex-valued U-Net} 
Auto-encoding~\cite{19} can effectively extract features by mapping input data to a latent space while preserving key information. Hence, we employ a complex-valued U-Net as an auto-encoder, optimizing it to approximate the input through its output. Given the complex-valued nature of k-space and the potential neighborhood relationships within itself, complex convolutions~\cite{20} are employed to preserve both phase and magnitude information while effectively aggregating contextual information from neighborhood. The complex-valued U-Net features a two-level architecture with a fixed channel number of 64. Firstly, two complex convolutional layers at both ends adjust the channel number. Then each encoder and decoder level incorporates a Complex Convolutional Recurrent Block (C-CRNN) to extract features from spatial and temporal neighborhoods, with recurrent operations applied in opposite temporal directions. Their outputs are concatenated and fused via a complex convolution to capture bidirectional temporal information while integrating low- and high-level frequency features. Similarly, a Complex Bidirectional Convolutional Recurrent Block (C-BCRNN) is incorporated at the bottleneck. As no downsampling is applied, k-space embeddings are directly extracted from decoder features using coordinate queries and fused via element-wise addition.


\subsubsection{P-INR \& K-INR Branches} Both K-INR and P-INR branches employ a 7-layer MLP. Following the suggestions from the INR tutorial\footnote{\url{https://github.com/INR4MICCAI/INRTutorial/tree/main}}, P-INR chooses to leverage an MLP network with complex Gabor wavelet activation (WIRE)~\cite{21} to process input positional embedding, enhancing its ability to handle sparse data while capturing both fine details and global structure effectively. To better recover high-frequency details and sharpen reconstructions, we separately embed spatial and temporal coordinates~\cite{40}, using NeRF's positional encoding~\cite{9} for the spatial component and Fourier positional encoding~\cite{31} for the temporal component. All intermediate layers produce a hidden feature dimension of 512. 

In the K-INR branch, an initial MLP adjusts the length of k-space feature embeddings, followed by another MLP with leaky ReLU activation function for effective feature processing, offering different information compared to WIRE. The output embedding length is initially 64 and expands to 512 after intermediate cross-branch interaction. 

To enhance interaction between two branches, we first apply a linear layer to each input embedding and add the result to the other branch’s input. Moreover, an MLP is placed in the middle part of two branches to process concatenated intermediate features. The fused output is concatenated back with each branch’s input and passed through the subsequent MLP network to generate the final prediction.


\subsubsection{Inference Stage}
In the inference stage, reconstructed multi-coil k-space is generated to guide subsequent optimization, as shown in Figure~\ref{fig1}. Given a Cartesian grid of all coordinates in the cine sequence, the outputs from both branches are averaged to produce the predicted multi-coil k-space. A hard data consistency (DC) operation is applied to preserve the originally sampled k-space values. The resulting multi-coil k-space is then used as input to the complex-valued U-Net in subsequent optimization stage. Furthermore, to obtain the final reconstruction, multi-coil k-space is transformed to the image domain using the Inverse Fast Fourier Transform (IFFT) and a coil combination step is performed to yield the final coil-combined image: the multi-coil images are multiplied by the conjugate of the coil sensitivity map (CSM) and summed across coils.

\subsection{Loss Function}
The loss function consists of four components: reconstruction losses $L_{\bm{y}_{kv}}$ and $L_{\bm{y}_{pv}}$ for the two branch outputs, $L_{ae}^{acq}$ for the auto-encoding of acquired k-space part, and $L_{ae}^{zf}$ for the auto-encoding of left zero-filled k-space part. Therefore, the overall loss is expressed as:
\begin{equation}
    L = \lambda_{1}L_{\bm{y}_{kv}}+\lambda_{2}L_{\bm{y}_{pv}}+\lambda_{3}L_{ae}^{acq}+\lambda_{4}L_{ae}^{zf}.
\end{equation}
Here, $\lambda_i$ are the weights for each loss component, with $\lambda_{1}=\lambda_{2}=1$,  $\lambda_{3}=0.05$ and $\lambda_{4}=0.025$, reflecting their relative importance. Since auto-encoding serves as an auxiliary task, its overall weight is lower. Additionally, auto-encoding of the acquired k-space is prioritized over the zero-filled part due to its grounding in actual measured data. $L_{\bm{y}_{kv}}$ and $L_{\bm{y}_{pv}}$ use the high dynamic range loss from NIK~\cite{14} to address imbalanced k-space values. The MSE loss is employed for two auto-encoding losses. Notably, each of them focuses solely on the corresponding k-space region by applying the mask and averaging the loss over that region.

\section{Experimental Setup}
\textbf{Dataset}. The proposed KP-INR was evaluated on the public CMRxRecon2024 dataset~\cite{24}, acquired using a 3T MAGNETOM Vida scanner (Siemens Healthineers) with dedicated multi-channel cardiac coils. The dataset includes 330 healthy subjects, covering multiple contrasts, anatomical views, and k-space undersampling trajectories. As this work focuses on cardiac cine MRI reconstruction, we primarily describe the cine data in long-axis (LAX) and short-axis (SAX) views, acquired with a TrueFISP sequence. The cine data includes 3 LAX slices and 8-14 SAX slices per subject, segmented into 12-25 cardiac phases with a temporal resolution of approximately 50 ms, spatial resolution of 1.5$\times$1.5 mm$^{2}$, slice thickness of 8.0 mm, and a 4.0 mm slice gap. More details are available in~\cite{24}. In this work, we randomly selected 9 subjects (4 LAX and 5 SAX) for optimization and evaluation. To comprehensively assess performance, we conducted experiments using two Cartesian sampling trajectories (uniform and Gaussian) and two acceleration factors (4$\times$, 8$\times$). Notably, all undersampling masks were in 3D format due to the application of temporal interleaving, with the central 16 lines fully sampled as the autocalibration signal (ACS) region.
\\
\\
\textbf{Implementation Details}. 
In KP-INR, the input coordinate set $P$ was normalized to $[-1, 1]$. NeRF's positional encoding~\cite{9} was applied to the spatial coordinate component $P_{s}$, formulated as $\gamma(P_{s}) = [\sin(2^{l}\pi P_{s}), \cos(2^{l}\pi P_{s})]_{l=0}^{L-1}$. Fourier positional encoding~\cite{31} was applied to the temporal coordinate component $P_{t}$ by generating a random Gaussian matrix $B$, sampled from a normal distribution $\mathcal{N}(0, 1)$, represented as $\gamma(P_{t}) = [\cos(2\pi BP_{t}), \sin(2\pi BP_{t})]$. Here, the spatial and temporal components yield embedding lengths of 480 and 96, respectively, resulting in a total input embedding length of 576.

We evaluated KP-INR against three baselines: L+S~\cite{32}, k-t GRAPPA~\cite{36}, and P-INR. L+S is a traditional method that reconstructs cine MRI by decomposing data into low-rank and sparse components. k-t GRAPPA enhances GRAPPA~\cite{1} by leveraging both spatial and temporal neighborhood relationships in k-space. P-INR follows the standard INR design (similar to NIK~\cite{14}) by learning the coordinate-value mappings, and also serves to isolate the contribution of the introduced K-INR branch.

The models were all implemented in PyTorch 2.0.0 on an NVIDIA A100 GPU with 40GB of GPU memory. INR methods were optimized using the AdamW optimizer with parameters $\beta_{1}=0.9$, $\beta_{2}=0.999$, $\epsilon=10^{-8}$, a learning rate of $5\times10^{-5}$, and weight decay 0.1. The learning rate was decreased by multiplying with 0.95 every 500 epochs, with a minimal value of $2\times10^{-6}$. Furthermore, the k-space input is iteratively updated every 500 epochs for the alternating optimization. The model was optimized for 6000 epochs to ensure sufficient convergence. For a comprehensive quantitative evaluation, peak signal-to-noise ratio (PSNR), structural similarity index (SSIM) and deep image structure and texture similarity (DISTS)~\cite{27} were computed. Notably, DISTS serves as an effective perceptual metric~\cite{28}. To avoid inflated evaluation metrics from non-informative black regions, we cropped the first and last quarters along the height dimension before calculating metrics, with additional cropping applied for visualization. Furthermore, the Wilcoxon signed-rank test was used to assess the statistical significance of the proposed method.

\section{Results}
\textbf{Comparison with Baseline Methods}. Table~\ref{tab1} reports the performance of all methods on the cropped region across two Cartesian sampling trajectories and acceleration factors. All baseline methods exhibit statistically significant differences compared to the proposed method. Under the uniform Cartesian sampling, both L+S and k-t GRAPPA perform comparably to P-INR at low acceleration factor, but their performance highly degrades at higher factors, revealing a limitation in scalability. For the Gaussian Cartesian sampling, a similar trend is observed, though the performance gap between INR methods and the L+S method narrows.



\begin{table}[!tb]
\caption{Comparison of KP-INR (proposed) with baseline methods. Best results are shown in bold. The $\dagger$ after each metric indicates a statistically significant difference (\textit{p} < .05) from the proposed method.}\label{tab1}
\centering
\resizebox{1\textwidth}{!}{
\begin{tabular}{c|c|l|c|c|c}
\hline
Sampling & R & Methods & PSNR$\uparrow$ & SSIM$\uparrow$ & DISTS$\uparrow$ \\ 
\hline
\multirow{8}{*}{\rotatebox{90}{Uniform Cartesian}} & \multirow{4}{*}{$4\times$} & L+S~\cite{32} & $38.60 \pm 3.01^\dagger$ & $0.9415\pm 0.0260^\dagger$ & $0.9344 \pm 0.022^\dagger$ \\
\multirow{3}{*}{} & \multirow{3}{*}{} & k-t GRAPPA~\cite{36} & $40.92 \pm 1.71^\dagger$ & $0.9491 \pm 0.0148^\dagger$ & $0.9376 \pm 0.0097^\dagger$  \\
\multirow{3}{*}{} & \multirow{3}{*}{} & P-INR & $39.96\pm 2.34^\dagger$ & $0.9668 \pm 0.0097^\dagger$ & $0.9276 \pm 0.0109^\dagger$  \\
\multirow{3}{*}{} & \multirow{3}{*}{} & KP-INR & $\textbf{41.99} \pm \textbf{2.18}$ & $\textbf{0.9748} \pm \textbf{0.0079}$ & $\textbf{0.9430} \pm \textbf{0.0094}$ \\
\cline{2-6}

\multirow{3}{*}{} & \multirow{4}{*}{$8\times$} & L+S~\cite{32} & $32.83 \pm 4.75^\dagger$ & $0.8497 \pm 0.1479^\dagger$ & $0.8661 \pm 0.0866^\dagger$  \\
\multirow{3}{*}{} & \multirow{3}{*}{} & k-t GRAPPA~\cite{36} & $29.24 \pm 1.76^\dagger$ & $0.7631 \pm 0.0551^\dagger$ & $0.8171 \pm 0.0193^\dagger$ \\
\multirow{3}{*}{} & \multirow{3}{*}{} & P-INR & $37.21 \pm 2.66^\dagger$ & $0.9425 \pm 0.0199^\dagger$ & $0.9042 \pm 0.0173^\dagger$ \\
\multirow{3}{*}{} & \multirow{3}{*}{} & KP-INR & $\textbf{37.85} \pm \textbf{2.48}$ & $\textbf{0.9502} \pm \textbf{0.0152}$ & $\textbf{0.9133} \pm \textbf{0.0151}$ \\
\hline
\hline

\multirow{8}{*}{\rotatebox{90}{Gaussian Cartesian}} & \multirow{4}{*}{$4\times$} & L+S~\cite{32} & $38.42 \pm 2.32^\dagger$ & $0.9343 \pm 0.0239^\dagger$ & $0.9294 \pm 0.0220^\dagger$ \\
\multirow{3}{*}{} & \multirow{3}{*}{} & k-t GRAPPA~\cite{36} & $39.18 \pm 1.97^\dagger$ & $0.9403 \pm 0.0177^\dagger$ & $0.9253 \pm 0.0105^\dagger$  \\
\multirow{3}{*}{} & \multirow{3}{*}{} & P-INR & $38.14 \pm 2.28^\dagger$ & $0.9455 \pm 0.0190^\dagger$ & $0.9128 \pm 0.0191^\dagger$  \\
\multirow{3}{*}{} & \multirow{3}{*}{} & KP-INR & $\textbf{39.98} \pm \textbf{2.24}$ & $\textbf{0.9649} \pm \textbf{0.0094}$ & $\textbf{0.9328} \pm \textbf{0.0091}$ \\
\cline{2-6}

\multirow{3}{*}{} & \multirow{4}{*}{$8\times$} & L+S~\cite{32} & $35.23 \pm 2.63^\dagger$ & $0.8901 \pm 0.0335^\dagger$ & $\textbf{0.8968} \pm \textbf{0.0244}^\dagger$  \\
\multirow{3}{*}{} & \multirow{3}{*}{} & k-t GRAPPA~\cite{36} & $33.04 \pm 2.04^\dagger$ & $0.8414 \pm 0.0476^\dagger$ & $0.8528 \pm 0.0238^\dagger$ \\
\multirow{3}{*}{} & \multirow{3}{*}{} & P-INR & $33.95 \pm 2.15^\dagger$ & $0.8703 \pm 0.0495^\dagger$ & $0.8510 \pm 0.0323^\dagger$ \\
\multirow{3}{*}{} & \multirow{3}{*}{} & KP-INR & $\textbf{35.86} \pm \textbf{2.01}$ & $\textbf{0.9123} \pm \textbf{0.0298}$ & $0.8861 \pm 0.0199$ \\
\hline

\end{tabular}
}
\end{table}

\begin{figure}[!tb]
\centering
\resizebox{1\textwidth}{!}{\includegraphics{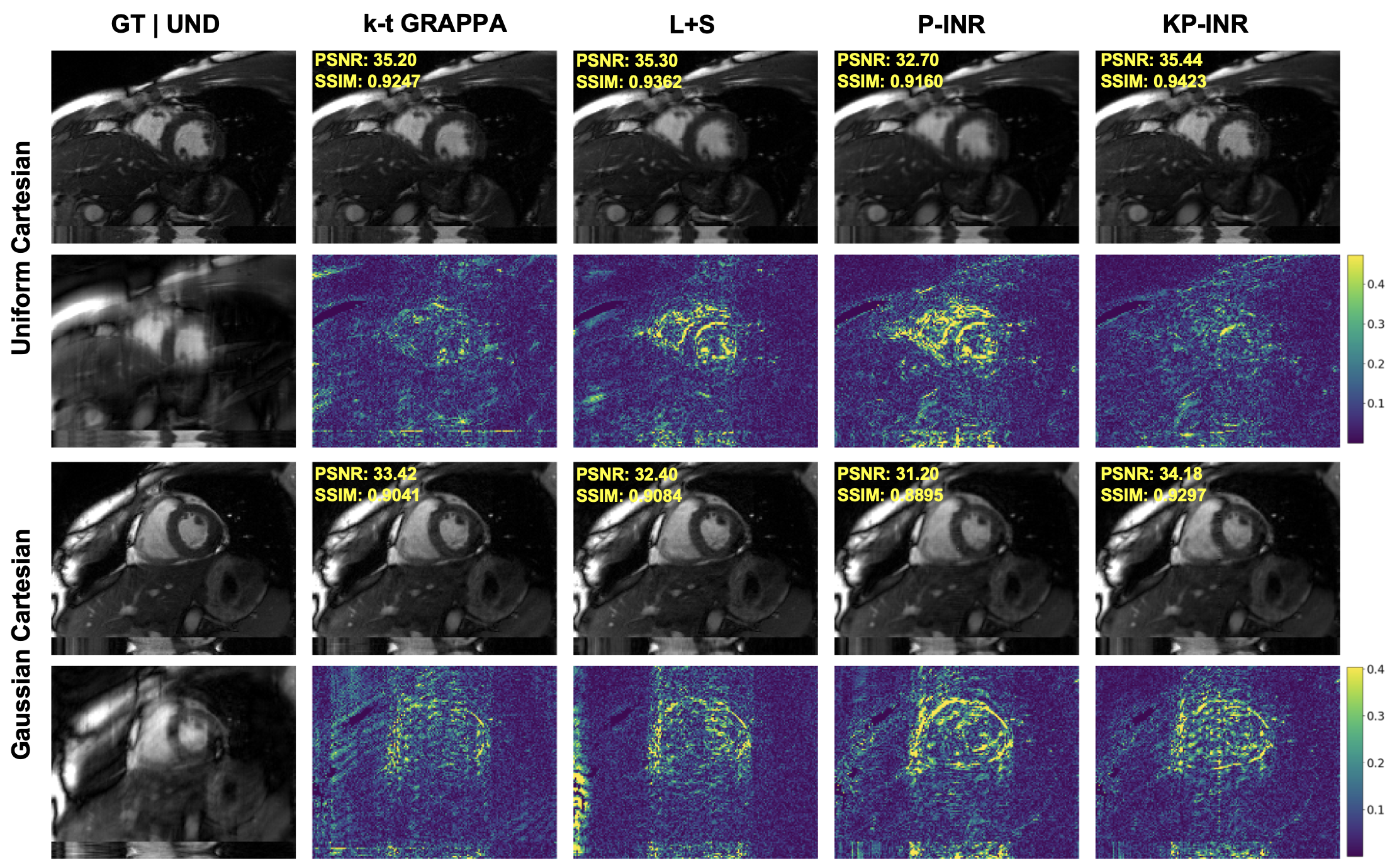}}
\caption{Visualization of reconstructed SAX samples under uniform Cartesian and Gaussian Cartesian sampling at $R=4$ is shown. The bottom section of each visualization presents the $x-t$ view reconstruction along the central line in the temporal dimension, with PSNR and SSIM values for the full cropped cine sequence displayed in the top-right corner.} \label{fig2}
\end{figure}

Figure~\ref{fig2} shows qualitative results under uniform Cartesian and Gaussian Cartesian trajectories at an acceleration factor of 4. Overall, KP-INR achieves the best reconstructions, effectively restoring fine details and removing aliasing artifacts. In contrast, P-INR struggles with Cartesian sampling patterns, exhibiting noticeable blurriness in critical cardiac regions. Unlike non-Cartesian sampling, Cartesian sampling lacks direction diversity, making it more challenging to learn the overall k-space distribution, which reflects the limitations of traditional INR design for Cartesian cine MRI. While k-t GRAPPA and L+S produce competitive visual results, they still exhibit residual aliasing and incomplete detail recovery, underscoring the advantages of KP-INR in this regard.

\section{Discussion and Conclusion}
As shown in Table~\ref{tab1} and Figure~\ref{fig2}, KP-INR outperformed an original INR design with only positional information, highlighting the benefits of incorporating k-space features in recovering finer details and sharper reconstructions. Similar to classical methods like k-t GRAPPA, our method leverages local structure in k-space, but here with convolutional operations that enrich representation with contextual information. Our alternating optimization strategy progressively refines k-space estimates and their latent representations, leading to improved performance. We also observe that uniform sampling outperforms Gaussian sampling pattern, likely due to the latter’s relative sparsity in high-frequency regions. Moreover, learning effective k-space feature representations and enabling interactions between two branches are critical for facilitating complementary learning. Further exploration of these two aspects offers a promising direction for enhancing the performance of KP-INR.
\\
\\
In this work, KP-INR extracts local structure of k-space and integrates these features with a standard positional INR, showing substantial improvement over P-INR and k-t GRAPPA. KP-INR therefore yields promise for further improving INR performance in cardiac cine MRI reconstruction.
\\
\\
\textbf{Acknowledgement.} This work is part of the project ROBUST: Trustworthy AI-based Systems for Sustainable Growth with
project number KICH3.LTP.20.006, which is (partly)
financed by the Dutch Research Council (NWO), Philips Research, and the Dutch Ministry of Economic Affairs
and Climate Policy (EZK) under the program LTP KIC
2020-2023.

\bibliographystyle{splncs04}
\bibliography{new}
\end{document}